\documentclass{article}

\usepackage{amsmath, amsthm, amssymb, amsfonts}
\usepackage{thmtools}
\usepackage{graphicx}
\usepackage{setspace}
\usepackage{geometry}
\usepackage{float}
\usepackage{hyperref}
\usepackage[utf8]{inputenc}
\usepackage[english]{babel}
\usepackage{framed}
\usepackage[dvipsnames]{xcolor}
\usepackage{tcolorbox}

\usepackage{algorithm} 
\usepackage{algpseudocode} 
\algnewcommand\algorithmicforeach{\textbf{for each}}
\algdef{S}[FOR]{ForEach}[1]{\algorithmicforeach\ #1\ \algorithmicdo}

\colorlet{LightGray}{White!90!Periwinkle}
\colorlet{LightOrange}{Orange!15}
\colorlet{LightGreen}{Green!15}

\newcommand{\HRule}[1]{\rule{\linewidth}{#1}}

\declaretheoremstyle[name=Theorem,]{thmsty}

\tcolorboxenvironment{theorem}{colback=LightGray}

\declaretheoremstyle[name=Proposition,]{prosty}

\tcolorboxenvironment{proposition}{colback=LightOrange}

\declaretheoremstyle[name=Principle,]{prcpsty}

\tcolorboxenvironment{principle}{colback=LightGreen}

\begin{document}

% ------------------------------------------------------------------------------
% Cover Page and ToC
% ------------------------------------------------------------------------------

\title{ \normalsize \textsc{}
		\\ [2.0cm]
		\HRule{1.5pt} \\
		\LARGE \textbf{\uppercase{Improved path planning algorithms for non-holonomic autonomous vehicles in industrial environments with narrow corridors: Roadmap Hybrid A* and Waypoints Hybrid A*}
		\HRule{2.0pt} \\ [0.6cm] \LARGE{Roadmap Hybrid A* and Waypoints Hybrid A* Pseudocodes} \vspace*{2\baselineskip}}
		}
\date{}

\author{\textbf{Alessandro Bonetti} \\ 
		University of Modena and Reggio Emilia, \\ Via Amendola 2, Pad. Morselli - 42122 Reggio Emilia, Italy \\
		alessandro.bonetti@unimore.it \\ \\
  \textbf{Simone Guidetti} \\ 
		Gruppo TecnoFerrari S.p.a. con socio unico,\\ Via Ghiarola Nuova, 105, 41042 Fiorano Modenese (MO), Italy \\
		simone\_guidetti@tecnoferrari.it \\ \\
  \textbf{Lorenzo Sabattini} \\ 
		University of Modena and Reggio Emilia, \\ Via Amendola 2, Pad. Morselli - 42122 Reggio Emilia, Italy \\
		lorenzo.sabattini@unimore.it
  
		}

\maketitle
\newpage

\tableofcontents
\newpage

% ------------------------------------------------------------------------------

\section{Proposed solutions} \label{sec:Proposed solution}
\begin{figure}[thb]%[H]
\centering
\includegraphics[width=12cm]{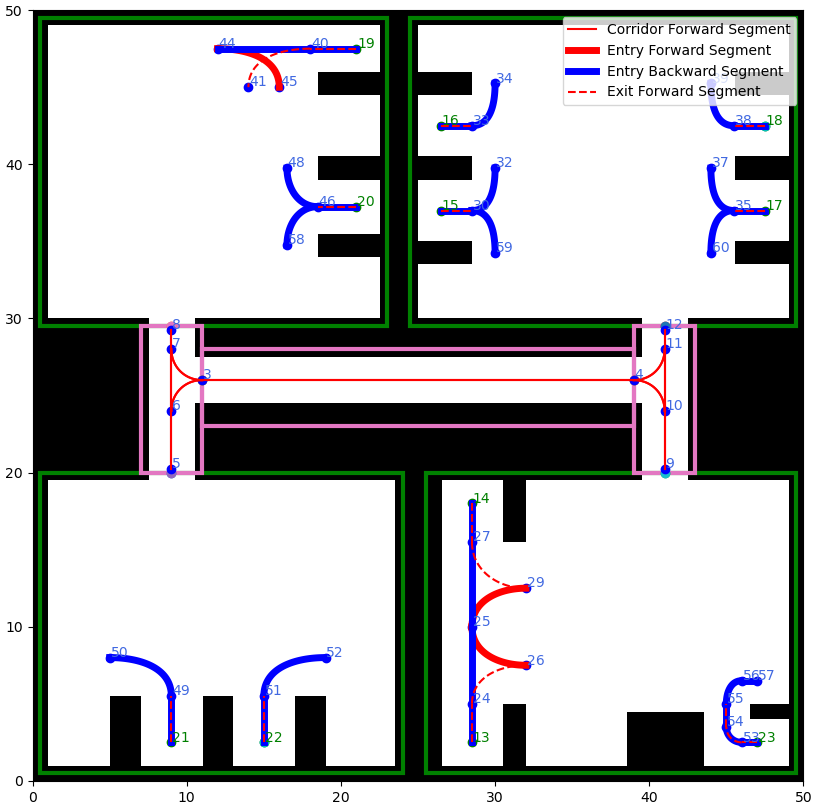}
\caption{The image shows the topological map of the plant, highlighting areas with green borders containing machines and narrow corridors marked in pink. The map features Bezier curve segments, with the red curves driven forward by the vehicle and the blue ones in reverse. Furthermore, the legend identifies the different types of segments used in corridors and entry and exit maneuvers. The numbers in the image uniquely identify the endpoints of each segment Bezier curve, enabling to define connections between them. Two segments are connected if they share the same endpoint.}
\label{fig2}
\end{figure}

In order to overcome the issues that have arisen from the standard version of Hybrid A* in the industrial environment described in Section 2, two new global path planners are presented: Roadmap Hybrid A* and Waypoint Hybrid A*.
For the development of both algorithms, some preliminary steps were required. To begin, the map was divided manually into rectangular zones. This division aimed to provide a topological representation of the environment, composed of machine servicing areas and corridors. The former are represented by green rectangles, while the corridors are depicted by pink rectangles, as shown in Fig.~\ref{fig2}. A topological graph of the plant was then set up using the NetworkX  \cite{SciPyProceedings_11} python library by imposing the connections between these rectangular zones.

To complete the preliminary steps required for implementing the algorithms, we manually designed the corridor segments and the machine entrance and exit segments for the AMR using Bezier curves. The corridor curves were drawn in the center to maximize the distance between the AMR and the walls, while machinery entry and exit curves were designed in order to minimize the vehicle footprint while maneuvering and ensuring safety. To accomplish this, we carefully selected the control points of each Bezier curve so as to obtain a collision-free and feasible path for the vehicle. The former condition is achieved by applying a collision checking algorithm that compares the bounding box of the vehicle with the grid map cells on the poses traversed by the vehicle. The latter condition is achieved by checking the maximum curvature of each curve segment and ensuring that consecutive segments have matching tangents to guarantee smoothness.

Then we collected the connectivity relationships between all the curves into a graph, on which a graph search algorithm is applied to extract the fixed path parts needed to form the final path of the proposed algorithms.

In Fig.~\ref{fig2}, the red segments represent the forward fixed sections of the path traveled by the vehicle, while the blue segments represent the reverse ones. The blue segments are only used for the entrance to the machines, as the uploading and downloading tools are located at the back of the vehicle.

In conclusion, the topological map graph and the Bezier curve segment graph are obtained in order to implement the Roadmap Hybrid A* and Waypoint Hybrid A* algorithms, as detailed in the following subsections.

\subsection{Roadmap Hybrid A*} \label{subsec:Roadmap}
In this subsection, we propose Roadmap Hybrid A*, a novel path planning technique for autonomous mobile vehicles subject to non-holonomic constraints. The technique combines two different methods: a graph search algorithm applied to fixed segments and the Hybrid A* algorithm. The former is used in obstacle-free zones and for maneuvering in and out of machines to guarantee a robust and predictable path. The latter provides flexibility and the ability to navigate around obstacles in more dynamic areas.

\def\Plus{\texttt{+}}

\begin{algorithm}[bth]
\caption{RoadMap Hybrid A*}
\hspace*{\algorithmicindent} \textbf{Input: $q_{start}, q_{goal}, TopologicalGraph, SegmentGraph, GridMap$} \\
\hspace*{\algorithmicindent} \textbf{Output: $FinalPath$}
\begin{algorithmic}[1]
\State $ExitPath, DetachmentPose \leftarrow FindExitPath(q_{start}, SegmentGraph)$
\State $EntryPath, AttachPose \leftarrow FindEntryPath(q_{goal}, SegmentGraph)$
\ForEach {$area\in TopologicalGraph$}
\If{$EvenOddRule(q_{start}, area) $}
\State $StartArea = area$
\EndIf
\If{$EvenOddRule(q_{goal}, area) $}
\State $GoalArea = area$
\EndIf
\EndFor
\If{$StartArea \neq GoalArea$}
\State $AreasSequence \leftarrow DijkstraAlgorithm(TopologicalGraph, StartArea, GoalArea) $
\State $CorridorsSequence \leftarrow FindCorridorsSequence(AreasSequence) $
\State $ CorridorPath, EndPoints \leftarrow FindCorridorsPath(CorridorsSequence, SegmentGraph) $
\State $HybridAstarPaths \leftarrow \emptyset$
\State $numEndPoints \leftarrow$ number of $EndPoints$
\For{$i=1$ to $numEndPoints$}
\If{$i = 1$}
\State $HybridAstarPath_i \leftarrow HybridAstar(DetachmentPose, EndPoints[i], GridMap)$
\ElsIf{$i = numEndPoints$}
\State $HybridAstarPath_i \leftarrow HybridAstar(EndPoints[i], AttachPose, GridMap)$
\ElsIf{$i$ is $even$}
\State $HybridAstarPath_i \leftarrow HybridAstar (EndPoints[i], EndPoints[i\Plus1], GridMap)$
\Else
\State continue to next iteration
\EndIf
\State $HybridAstarPaths \leftarrow HybridAstarPaths \bigcup HybridAstarPath_i$
\EndFor
\Else
\State $HybridAstarPaths \leftarrow HybridAstar(DetachmentPose, AttachPose, GridMap)$
\EndIf
%\State $HybridAstarPaths \leftarrow \bigcup\limits_{i=1}^{n} HybridAstarPath_i$
\State $FinalPath \leftarrow ExitPath\bigcup CorridorPath \bigcup HybridAstarPaths\bigcup EntryPath$
\State \textbf{return} $FinalPath$
\end{algorithmic}
\label{alg2}
\end{algorithm}

Roadmap Hybrid A*, as described %shown 
in pseudo-code in Algorithm~\ref{alg2}, uses the start and goal nodes of the vehicle, the Bezier segment graph, the topological graph, and the grid map as input to initiate the planning process. The first step is to identify the entry and exit paths, as well as the corresponding attachment and detachment nodes. The exit path is the fixed segment curve that allows the vehicle to safely exit the starting machine and whose end points are the start and detachment nodes. Following the same principle, the entry path is the fixed curve that the AMR must travel to enter the destination station and whose end points are the goal and detachment nodes. The attachment and detachment nodes serve as transition points between the free motion of the vehicle provided by Hybrid A* and the fixed path during the entry or exit processes.

The next step involves determining the start and goal areas using the Even-Odd rule~\cite{FoleyDamEtAl90}, which is a technique used in computational geometry to determine if a point lies inside or outside a closed polygonal curve in a two-dimensional plane. The rule consists in drawing a straight line from that point in any direction and counting the number of intersections of the line with the shape. If this number is odd, the point is inside; if even, the point is outside. The Even-Odd rule works for any simple or complex polygon, and it was applied to the rectangular areas and the positions of the start and goal machine nodes, as shown from line 3 to 10 of the pseudo-code. If the start and goal areas are different, Roadmap Hybrid A* uses Dijkstra's algorithm to determine the sequence of areas that the vehicle must traverse. The sequence of corridors, the Bezier curve paths within them and the corresponding ordered endpoint list are also found at this stage. As shown from lines 17 to 28 of Algorithm~\ref{alg2}, the fixed-path segment endpoints, as well as the attachment and detachment nodes, are then connected using standard Hybrid A*. The detachment pose is linked to the first endpoint, and the even-indexed endpoints are linked to the odd-indexed ones. Finally,  the last endpoint is connected to the attachment node. As shown in line 30 of Algorithm~\ref{alg2}, if the initial area matches the target area instead, the detachment pose is directly connected with the attachment pose using Hybrid A*.

\begin{figure}[bth]%[H]
\centering
\includegraphics[width=12cm]{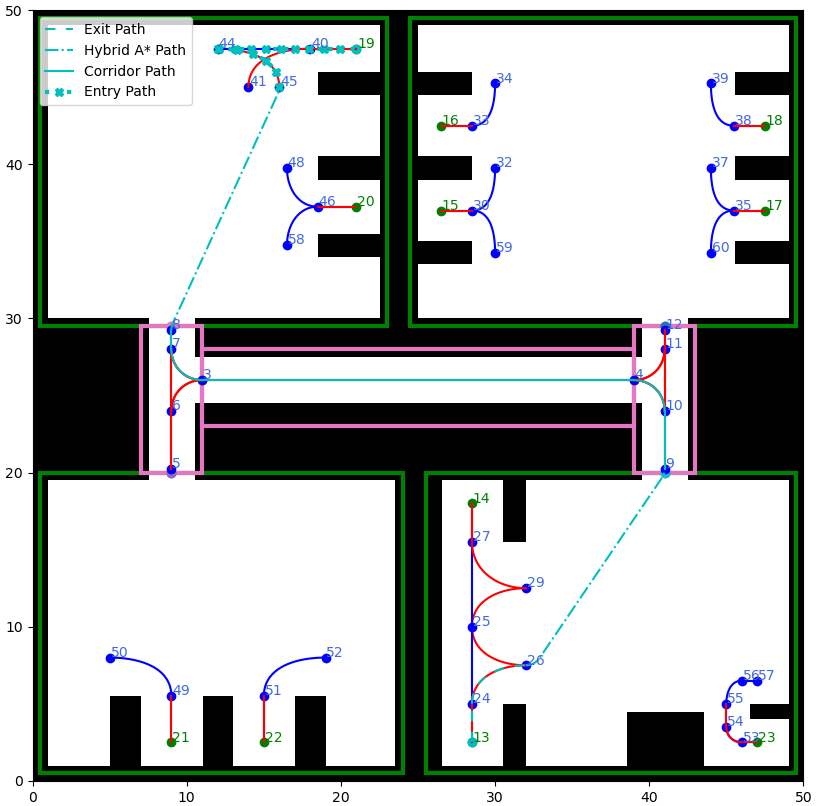}
\caption{Roadmap Hybrid A* path planned from node 13 to 19. The legend identifies the exit path, Hybrid A* path, corridor path and entry path parts.}
\label{fig3}
\end{figure}

Finally, the exit path, corridor path, Hybrid A* paths, and entry path are concatenated to produce the final output of the algorithm. An example of RoadMap Hybrid A* is shown in Fig. \ref{fig3}.

%pseudocodice

%Sedighi, Nguyen, and Kuhnert discovered that
\newpage
\subsection{Waypoint Hybrid A*}
In this subsection, the Waypoint Hybrid A* path planner is presented. The implementation of this algorithm aimed to evaluate the cost-effectiveness of using waypoints in narrow corridors compared to the static roadmap employed in Roadmap Hybrid A*.

Waypoint Hybrid A* takes inspiration from the planner described in~\cite{8813752}, where waypoints were generated by applying a visibility graph and then connected by means of Hybrid A*. In \cite{8813752}, it has been found that using waypoints to guide the Hybrid A* to its destination results in a 40\% faster run-time. In this research we propose to adapt the waypoints principle in a slightly different way in order to speed up computational time but also trying to avoid oscillating paths produced by Hybrid A*.

\begin{algorithm}[bth]
\caption{Waypoint Hybrid A*}
\hspace*{\algorithmicindent} \textbf{Input: $q_{start}, q_{goal}, TopologicalGraph, SegmentGraph, GridMap$} \\
\hspace*{\algorithmicindent} \textbf{Output: $FinalPath$}
\begin{algorithmic}[1]
\State $ExitPath, DetachmentPose \leftarrow FindExitPath(q_{start}, SegmentGraph)$
\State $EntryPath, AttachPose \leftarrow FindEntryPath(q_{goal}, SegmentGraph)$
\ForEach {$area\in TopologicalGraph$}
\If{$EvenOddRule(q_{start}, area) $}
\State $StartArea = area$
\EndIf
\If{$EvenOddRule(q_{goal}, area) $}
\State $GoalArea = area$
\EndIf
\EndFor
\If{$StartArea \neq GoalArea$}
\State $ AreasSequence \leftarrow DijkstraAlgorithm(TopologicalGraph, StartArea, GoalArea) $
\State $ Waypoints \leftarrow FindWaypointsSequence(AreasSequence) $
%\State $EndingPoints \leftarrow FindEndingPoints(CorridorsSequence) $
\State $HybridAstarPaths \leftarrow \emptyset$
\State $numWaypoints \leftarrow$ number of $Waypoints$
\For{$i=1$ to $numWaypoints$}
\If{$i = 1$}
\State $HybridAstarPath_0 \leftarrow HybridAstar(DetachmentPose, Waypoints[i], GridMap)$
\State $HybridAstarPaths \leftarrow HybridAstarPaths \bigcup HybridAstarPath_0$
\State $HybridAstarPath_{i} \leftarrow HybridAstar(Waypoints[i], Waypoints[i\Plus1], GridMap)$
\ElsIf{$i = numWaypoints$}
\State $HybridAstarPath_i \leftarrow HybridAstar(Waypoints[i], AttachPose, GridMap)$
\Else
\State $HybridAstarPath_i \leftarrow HybridAstar (Waypoints[i], Waypoints[i\Plus1], GridMap)$
\EndIf
\State $HybridAstarPaths \leftarrow HybridAstarPaths \bigcup HybridAstarPath_i$
\EndFor
\Else
\State $HybridAstarPaths \leftarrow HybridAstar(DetachmentPose, AttachPose, GridMap)$
\EndIf

\State $FinalPath \leftarrow ExitPath\bigcup HybridAstarPaths\bigcup EntryPath$
\State \textbf{return} $FinalPath$
\end{algorithmic}
\label{alg3}
\end{algorithm}

Waypoint Hybrid A*, as presented in the pseudo-code in Algorithm~\ref{alg3}, uses the start and goal nodes of the vehicle, the Bezier segment graph, the topology graph, and the grid map of the environment as input. To begin the planning process, as explained with Roadmap Hybrid A* in Subsection~\ref{subsec:Roadmap}, the algorithm finds the entry and exit paths as well as the attachment and detachment nodes. After that, from lines 3 to 10, the initial and target areas are determined by applying the Even-Odd rule to the rectangular shape zones and the start and goal node positions. If the starting and target areas are different, the algorithm employs Dijkstra's algorithm to determine the sequence of areas through which the vehicle must pass. The ordered sequence of waypoints is then defined as the midpoint of the free space connection width between consecutive zones.

\begin{figure}[tbh]%[H]
\centering
\includegraphics[width=12cm]{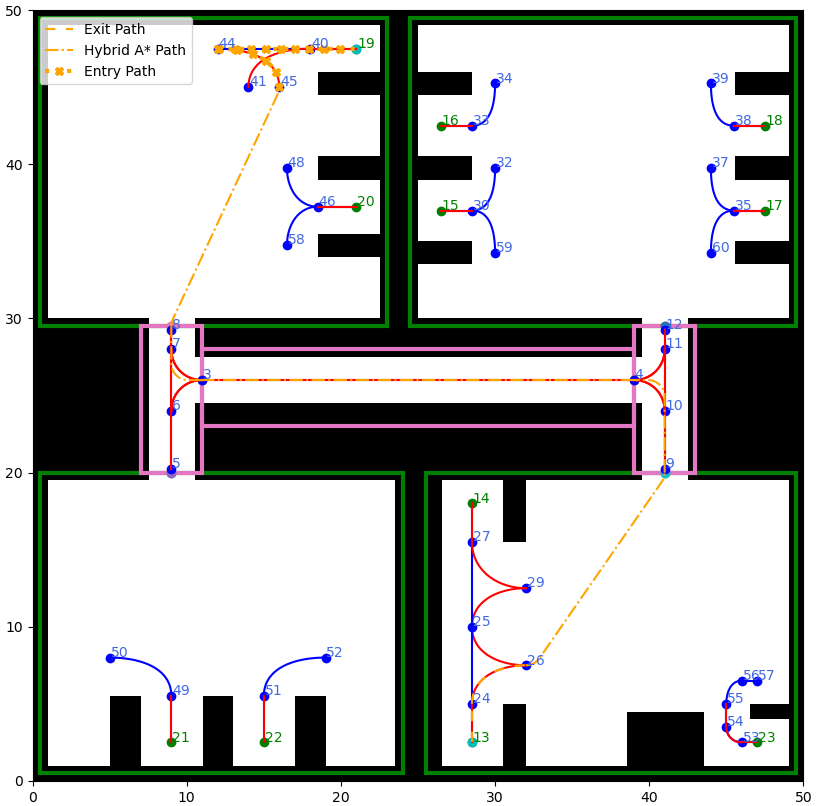}
\caption{Waypoint Hybrid A star path planned from node 13 to 19. The legend identifies the exit path, Hybrid A* path and entry path parts.}
\label{fig4}
\end{figure}

From line 16 to 27 of Algorithm~\ref{alg3}, the waypoints, the attachment node and detachment node are connected using standard Hybrid A*. The detachment pose is connected to the first waypoint, then each waypoint is connected with its successor except for the last one, which is connected with the last endpoint. As shown in line 29 of Algorithm~\ref{alg3}, if the initial area matches the target area, the detachment pose is directly connected with the attachment pose using Hybrid A*. It is worth noting that, if this condition is verified, the output of Roadmap Hybrid A* and Waypoint Hybrid A* are exactly the same. Finally, the exit path, Hybrid A* paths, and entry path are concatenated to produce the final path of the algorithm, as show in Fig.~\ref{fig4}.

\newpage

% ------------------------------------------------------------------------------
% Reference and Cited Works
% ------------------------------------------------------------------------------

\bibliographystyle{IEEEtran}
\bibliography{References.bib}

\begin{thebibliography}{1}
\providecommand{\url}[1]{#1}
\csname url@samestyle\endcsname
\providecommand{\newblock}{\relax}
\providecommand{\bibinfo}[2]{#2}
\providecommand{\BIBentrySTDinterwordspacing}{\spaceskip=0pt\relax}
\providecommand{\BIBentryALTinterwordstretchfactor}{4}
\providecommand{\BIBentryALTinterwordspacing}{\spaceskip=\fontdimen2\font plus
\BIBentryALTinterwordstretchfactor\fontdimen3\font minus
  \fontdimen4\font\relax}
\providecommand{\BIBforeignlanguage}[2]{{%
\expandafter\ifx\csname l@#1\endcsname\relax
\typeout{** WARNING: IEEEtran.bst: No hyphenation pattern has been}%
\typeout{** loaded for the language `#1'. Using the pattern for}%
\typeout{** the default language instead.}%
\else
\language=\csname l@#1\endcsname
\fi
#2}}
\providecommand{\BIBdecl}{\relax}
\BIBdecl

\bibitem{SciPyProceedings_11}
A.~A. Hagberg, D.~A. Schult, and P.~J. Swart, ``Exploring network structure,
  dynamics, and function using networkx,'' in \emph{Proceedings of the 7th
  Python in Science Conference}, G.~Varoquaux, T.~Vaught, and J.~Millman, Eds.,
  Pasadena, CA USA, 2008, pp. 11 -- 15.

\bibitem{FoleyDamEtAl90}
J.~D. Foley, A.~van Dam, S.~Feiner, and J.~Hughes, \emph{Computer Graphics:
  Principles and Practice}.\hskip 1em plus 0.5em minus 0.4em\relax Reading, MA:
  Addison-Wesley, 1990.

\bibitem{8813752}
S.~Sedighi, D.-V. Nguyen, and K.-D. Kuhnert, ``Guided hybrid a-star path
  planning algorithm for valet parking applications,'' in \emph{2019 5th
  International Conference on Control, Automation and Robotics (ICCAR)}, 2019,
  pp. 570--575.

\end{thebibliography}

% ------------------------------------------------------------------------------

\end{document}